\documentclass[letterpaper, 10 pt, conference]{ieeeconf}

\IEEEoverridecommandlockouts
\usepackage{cite}

\usepackage{graphicx}
\graphicspath{{./figures/}}

\usepackage{siunitx}
\usepackage{textcomp}
\usepackage{amsmath}
\usepackage{amssymb}
\usepackage{multirow}
\usepackage{subfigure}
\usepackage{color}
\usepackage{float}
\title{\LARGE \bf
Targetless Extrinsic Calibration of Stereo Cameras, Thermal Cameras, and Laser Sensors in the Wild
}

\author{Taimeng Fu, Huai Yu, Wen Yang, Yaoyu Hu and Sebastian Scherer
\thanks{This work has been submitted to the IEEE for possible publication. Copyright may be transferred without notice, after which this version may no longer be accessible.}
\thanks{Taimeng Fu is with the School of Electronic Information, Wuhan University, Wuhan 430072, China, and also with the School of Data Science,
        The Chinese University of Hong Kong, Shenzhen, 2001 Longxiang Boulevard, Longgang District, Shenzhen, China, 518712.
        E-mail: \tt\small taimengfu@link.cuhk.edu.cn}%
\thanks{Huai Yu and Wen Yang are with the School of Electronic Information, Wuhan University,  Wuhan 430072, China. E-mail: \tt\small \{yuhuai,yangwen\}@whu.edu.cn}
\thanks{Yaoyu Hu and Sebastian Scherer are with the AirLab, Carnagie Mellon University, Pittsburgh, PA 15213
        E-mail: {\tt\small \{yaoyuh, basti\}@andrew.cmu.edu}}
}

\begin{document}

\maketitle
\thispagestyle{empty}
\pagestyle{empty}


\begin{abstract}

The fusion of multi-modal sensors has become increasingly popular in autonomous driving and intelligent robots since it can provide richer information than any single sensor, enhance reliability in complex environments. Multi-sensor extrinsic calibration is one of the key factors of sensor fusion. However, such calibration is difficult due to the variety of sensor modalities and the requirement of calibration targets and human labor. In this paper, we demonstrate a new targetless cross-modal calibration framework by focusing on the extrinsic transformations among stereo cameras, thermal cameras, and laser sensors. Specifically, the calibration between stereo and laser is conducted in 3D space by minimizing the registration error, while the thermal extrinsic to the other two sensors is estimated by optimizing the alignment of the edge features. Our method requires no dedicated targets and performs the multi-sensor calibration in a single shot without human interaction. Experimental results show that the calibration framework is accurate and applicable in general scenes.



\end{abstract}


\section{INTRODUCTION}
Extrinsic calibration between multi-modal sensors is a fundamental task for a variety of computer vision and robot applications, such as visual sensor fusion \cite{yu2020line} and object fusion \cite{zhang2019object}, 3D reconstruction \cite{cao2018depth}, and SLAM \cite{beauvisage2020multimodal}. The calibration process is to estimate the rigid-body transformation between each two sensors, which serves as a bridge to connect sensors, providing fusion information for the perception system, and enhancing the robot's ability to perceive the environment. Over the past several years, substantial works have been proposed to solve this problem either by manual calibration or using specific calibration targets, such as checkerboard and AprilTag. However, these methods are suffering from limited flexibility and are inaccurate on dynamic systems in which the extrinsic may slightly change after the calibration. In this paper, we mainly focus on the automatic extrinsic calibration between stereo cameras, thermal cameras, and laser sensors in the wild without any specific target. It has the great potential to overcome these limitations of existing calibration methods. 

\begin{figure}[htbp]
    \centering
    \includegraphics[width=\linewidth]{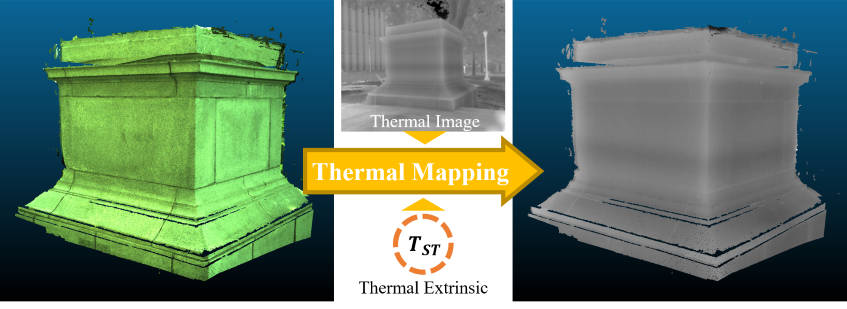}
    \caption{Thermal mapping on a dense reconstructed point cloud with the thermal extrinsic calibrated by our method.}
    \label{fig:recolor}
\end{figure}

We will review the recent works on the calibration of monocular/stereo visual cameras to laser sensors and visual to thermal cameras, then propose our contributions beyond these works.

\subsection{Calibration of monocular camera and laser sensor}
Existing literature on the calibration of monocular cameras and laser sensors can be classified according to 1) whether they require a specific calibration target (e.g. checkerboard), 2) whether they require manually pairing the laser points and camera pixels. Scaramuzza \emph{et al.} \cite{c2} demonstrated a laser point cloud visualization technique that enables manually labeling the correspondences between laser points and image pixels. They employed the perspective-from-n-points (PnP) algorithm to calculate the transformation based on these matches. Their method does not depend on fixed calibration targets, however, the manual matching is time-consuming and inflexible. On the other side, Nunez \emph{et al.} \cite{c3} chose to increase the degree of automation with a fixed checkerboard. They developed an algorithm to automatically detect the checkerboard in laser's view and align it to camera image to get the extrinsic parameters. To improve the accuracy, multiple laser frames were aggregated with the help of an inertial measurement unit (IMU). Although their calibration process does not require much manual intervention, the usage of checkerboard still limits its flexibility, since the specific calibration scene is not always available.

There are also some attempts to get rid of both fixed scene and manual manipulation. Pandey \emph{et al.} \cite{c4} provided a solution of automatic targetless extrinsic calibration by maximizing mutual information (MI) of laser reflectivity and image intensity. They found that after aggregating 10 scans, their MI cost function becomes convex and thus easy to optimize. However, the laser reflectivity and image intensity might not be strongly correlated in some scenarios. This might result in inaccurate calibrations. Levinson and Thrun \cite{c5} developed an online system that automatically aligns laser edges to image edges to correct sensor drifting. They projected laser edge points onto the generated cost maps to get the edge alignment cost and tried to adjust the extrinsic parameters to lower it. However, since the laser scans are relatively sparse, some of the laser edge points may not precisely lie on the boundary of the objects, which affects the calibration accuracy. Besides, they used a greedy approach to reduce the cost, which is less efficient than a gradient-based optimizer.


\subsection{Calibration of stereo cameras and laser sensor}

The stereo-laser calibration is more straightforward than that of monocular cameras and lasers since both stereo cameras and lasers have depth information. Guindel \emph{et al.} \cite{c6} calibrated stereo cameras and lasers based on a four-hole calibration board. They developed a segmentation pipeline for extracting boundaries of the holes from stereo point cloud and laser point cloud respectively and calculated the extrinsic parameters by minimizing the registration distance between the clustered centroids of the four holes. Dhall \emph{et al.} \cite{c7} reported a method that relies on a calibration board with a visual tag on it. They obtained the 3D position of the board corners in both stereo and laser’s view, and solve a set of equations to get the transformation that minimizes the distance between corresponding corner points. However, these methods still need known calibration targets to formulate the easy feature correspondences, which have the same problem of limited flexibility.

\subsection{Calibration of visual and thermal cameras}
In recent years, with the development and wide use of thermal cameras, visual-thermal calibration becomes an unavoidable task for image fusion and object fusion \cite{yu2020line,zhang2019object}. Most of the attempts tried to build calibration targets that are distinguishable in both visual and thermal views. Li \emph{et al.} \cite{c8} placed LED bulbs on their calibration board. The bulbs have high intensity on both RGB and thermal images since they emit both light and heat. They developed an algorithm to localize the bulbs on images and calibrate the extrinsic parameters by minimizing the reprojection error. Shivakumar \emph{et al.} \cite{c9} mounted aluminum squares on black acrylic background to form a checkerboard. Since the checkerboard blocks have different color and thermal reflectivity, it is easy to recognize the pattern on both RGB and thermal images. They then employed OpenCV’s \cite{cit:opencv} camera calibration toolbox to estimate the extrinsic parameters.

\subsection{Our contribution framework}
Most of the calibration methods mentioned above require specific calibration targets. This configuration limits their flexibility and may results in inaccuracy on dynamic systems with unpredictable extrinsic drifts after the calibration. On the other side, the scene-independent ones either require manual matching or are not robust enough. Besides, all of them only focus on two-modal calibration. It’s inconvenient to use them sequentially when types of sensors increase.

To solve these problems, we propose an automatic, targetless, all-in-one calibration framework for stereo cameras, thermal cameras, and laser sensors in the wild. For the stereo-laser calibration, we develop a multi-frame ICP to register stereo and laser point clouds. To calibrate the thermal extrinsic, we optimize the alignment of the edge features in stereo and laser point clouds and thermal images. Since the edge features are commonly available, our method works in arbitrary environments. Besides, it calibrates the three sensors with a single round of data collection. This greatly simplifies the cross-modal multi-sensor calibration process.

\section{METHODOLOGY}

The goal of our algorithm is to take a series of $n$ synchronized stereo image pairs $ I^{left}_{1:n}, I^{right}_{1:n} $, thermal images $I^{thermal}_{1:n}$, and laser point clouds $C^{laser}_{1:n}$, captured in arbitrary scenes, and automatically optimize the initial guess of the 6-DoF rigid-body transformations to get accurate calibrations. The transformation is defined by six parameters $ \xi=\{ x, y, z, row, pitch, yaw \} $, where $ x, y, z $ are translations, and $ roll, pitch, yaw $ are Euler angle rotations. We take the stereo left camera's coordinate system $\hat{S}$ as the base coordinate system, and calibrate other sensors to $\hat{S}$. As the stereo right to left $4\times4$ transformation matrix $T_{LR}$ can be easily obtained with tools such as OpenCV \cite{cit:opencv}, there are two transformations remaining to be estimated: laser to stereo transformation matrix $T_{SL}$, and thermal to stereo transformation matrix $T_{ST}$. We assume that the stereo and thermal cameras' lens distortions have been calibrated such that the pinhole camera model is applicable, and let $K^{left}, K^{right},$ and $K^{thermal}$ be the intrinsic matrices of these cameras, respectively.

\begin{figure}[htbp]
    \centering
    \includegraphics[width=\linewidth]{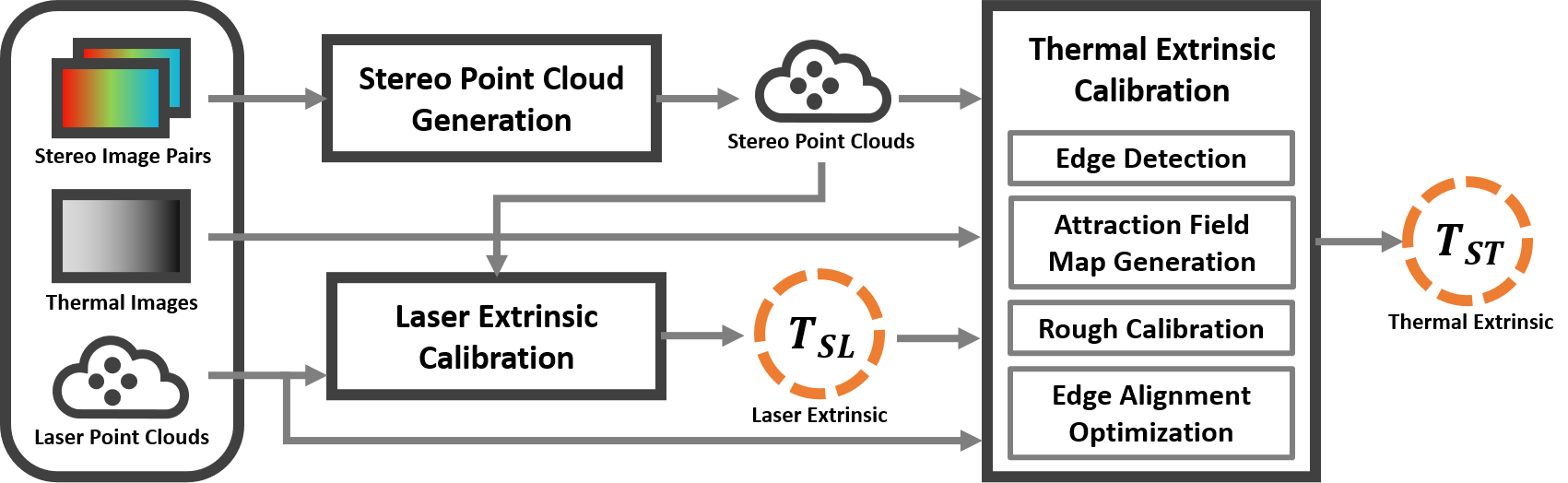}
    \caption{Flow chart of our calibration framework. It takes synchronized stereo image pairs, thermal images, and laser point clouds as input, and automatically calibrates the laser-stereo and thermal-stereo transformations in one system.}
    \label{fig:process}
\end{figure}

Our calibration framework has three steps. First, it generates stereo point clouds by image feature matching and triangulation. Then, we calibrate the stereo and laser by point cloud registration. Finally, our model optimizes the thermal-stereo transformation by minimizing the alignment error of the edge features detected in the three sensors. The flow of the system is shown in Fig.~\ref{fig:process}.

\subsection{Stereo point cloud generation} \label{stereosfm}

\begin{figure}[htbp]
    \centering
    \includegraphics[width=\linewidth]{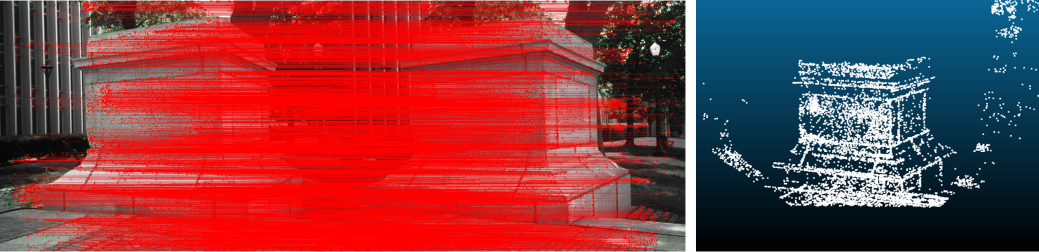}
    \caption{Example of the matched feature points in a stereo image pair (left) and the generated stereo point cloud (right).}
    \label{fig:stereo_match_pc}
\end{figure}

We generate point clouds from stereo image pairs to make full use of their 3D information in later calibrations. The Scale Invariant Feature Transform (SIFT) feature detection algorithm \cite{c12} is employed to extract key points and compute descriptors on $n$ stereo image pairs $(I^{left}_i, I^{right}_i), i=1 \cdots n$. A RegionsMatcher implemented in OpenMVG \cite{c50} is utilized to conduct the feature matching. Then the matched features are triangulated to get the 3D positions $p^{stereo}_{i,j}$, where $j$ is the index of matched features in image pair $i$. We use the OpenCV’s \cite{cit:opencv} triangulation function here. The stereo point clouds $C^{stereo}_i=\{p^{stereo}_{i, j}\}, i=1 \cdots n$. An example of stereo point cloud generation is shown in Fig. \ref{fig:stereo_match_pc}.


\subsection{Stereo-laser calibration} \label{lasercalib}

The aim of this part is to estimate the laser to stereo transformation $ T_{SL} $. We use the Iterative Closest Point (ICP) algorithm \cite{c10} to register the stereo and laser point clouds. We modify the original algorithm to a multi-frame ICP (MFICP) that considers point clouds from all frames in the dataset to optimize a unified extrinsic. It takes an initial guess of the laser extrinsic and iteratively optimizes the initial guess to minimize the distance between the closest points in each corresponding point cloud pair $(C^{stereo}_i, C^{laser}_i), i=1 \cdots n$, where $n$ is the number of frames in the dataset. Specifically, in each iteration, suppose $p^{i}_{k_1} \in C^{stereo}_i$, let $q^{i}_{k_2}$ be the closest point to $T^{-1}_{SL} p^{i}_{k_1}$ in $C^{laser}_i$, the registration cost is formulated as

\begin{equation}
Cost_{MFICP} = \sum_{i=1}^n \sum_{k_1, k_2} \left\| T^{-1}_{SL} p^{i}_{k_1} - q^{i}_{k_2} \right\|^{2}_{2}
\end{equation}

As a standard nonlinear optimization problem, we use Ceres solver \cite{cit:ceres} to optimize the initial guess of $T_{SL}$.


\subsection{Thermal extrinsic calibration}

Visual-thermal and laser-thermal calibrations are challenging because of the appearance difference and modality gap \cite{yu2020line}. Therefore, finding common features in RGB images, thermal images, and laser point clouds is the key to build connections. We observed that although the interior regions of textured object surfaces are not very distinguishable in thermal images, the object shape and edges are usually clear. Besides, the object outlines are also distinguishable in RGB and laser data. This leads us to the solution of calibrating the thermal extrinsic by aligning the edges in RGB images, thermal images, and laser point clouds.


\subsubsection{Stereo edge points detection}

\begin{figure}[htbp]
    \centering
    \includegraphics[width=\linewidth]{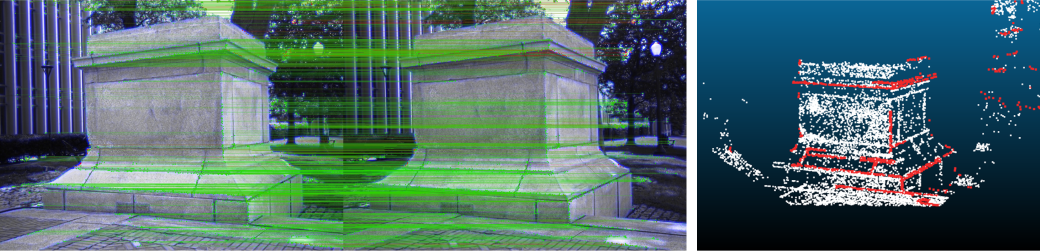}
    \caption{Example of the matched features on edges in a stereo image pair (left) and the edge points (marked in red) detected in the stereo point cloud (right).}
    \label{fig:stereo_edge_pc}
\end{figure}

Detecting stereo edge points is straightforward since we can first detect edges on stereo images and then triangulate the matched 2D edge points to 3D space (see Section \ref{stereosfm}). We use Sobel operator \cite{c13} to detect edges on each stereo image pair $(I^{left}_i, I^{right}_i), i=1 \cdots n$, then pick out the image feature points on these edges, and mark their corresponding 3D points as stereo edge points. Fig.~\ref{fig:stereo_edge_pc} gives one example of the stereo edge point detection. The edge points in the $i^{th}$ stereo cloud forms $E^{stereo}_i$.

\subsubsection{Laser edge points detection}

\begin{figure}[htbp]
    \centering
    \includegraphics[width=\linewidth]{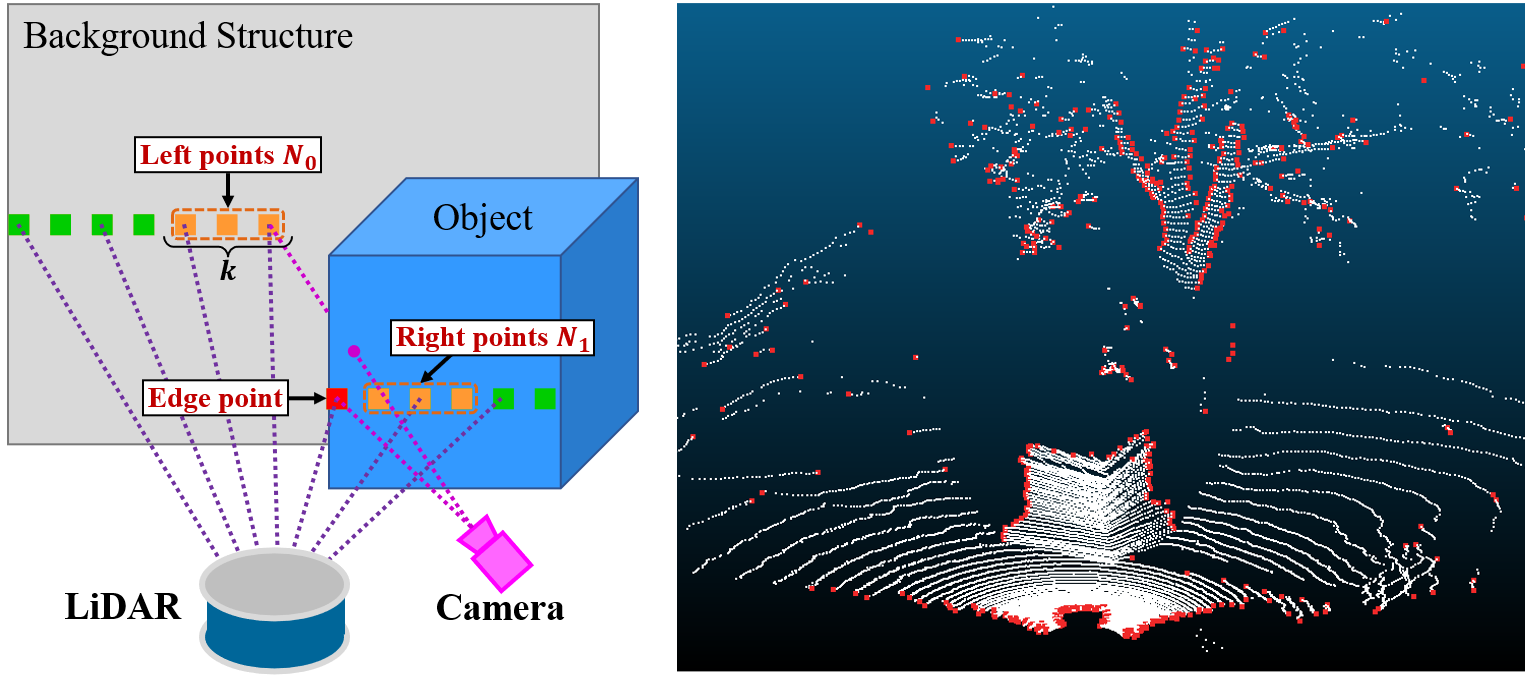}
    \caption{\emph{Left:} Schematic diagram of laser edge points detection with sampling radius $k=3$. \emph{Right:} The detected edge points (marked in red) in a laser point cloud.}
    \label{fig:laser_edge}
\end{figure}

Because the laser point density along the perpendicular direction of the scanning lines is too small, we only use the horizontal depth difference between adjacent scans to detect laser edges. For a laser point $p$, we collect its $k$ relative left points and $k$ relative right points, written as $N_0$ and $N_1$, as shown in Fig.~\ref{fig:laser_edge} (left). Then we define binary states $\alpha_{\lambda}$ as whether all points in $N_{\lambda}$ is close to $p$, and $\beta_{\lambda}$ as whether all points in $N_{\lambda}$ is deeper than $p$,
\begin{equation}
\begin{aligned}
\alpha_{\lambda} &= \forall p^{\prime} \in N_{\lambda}, \left|\|p^{\prime}\|_2 - \|p\|_2\right| \leq \epsilon \\
\beta_{\lambda} &= \forall p^{\prime} \in N_{\lambda}, \|p^{\prime}\|_2 - \|p\|_2 > \epsilon, &\lambda=0,1
\end{aligned}
\end{equation}
where $\epsilon$ is the depth difference threshold. Then the laser edge points are picked by
\begin{equation}
\begin{aligned}
    p\text{ is edge point} \Leftarrow \left( \alpha_0 \land \beta_1 \right) \lor \left( \beta_0 \land \alpha_1 \right)
\end{aligned}
\end{equation}
This picks the points with smaller depth at laser depth discontinuity edges, which will still be visible when viewed from the camera (while the deeper ones may not), as shown in Fig.~\ref{fig:laser_edge} (left). An example of the laser edge point detection result is given in Fig.~\ref{fig:laser_edge} (right). All edge points in the $i^{th}$ laser frame forms $E^{laser}_i$.

\subsubsection{Thermal edge detection and attraction field map generation} \label{afm}

\begin{figure}[htbp]
    \centering
    \includegraphics[width=\linewidth]{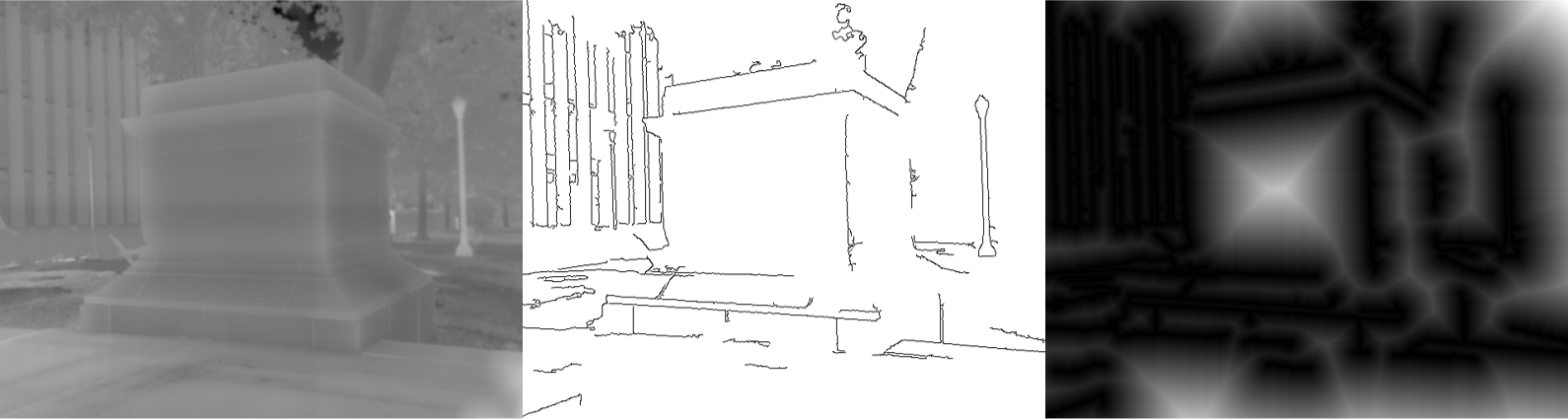}
    \caption{Example of a thermal image (left), its edge map (middle), and its attraction field map (right). Each pixel value in the attraction field map represents the normalized L2 distance from itself to the nearest edge.}
    \label{fig:thermal_edge_trac}
\end{figure}

We use the Canny edge detector \cite{c14} to detect edges on thermal images $I^{thermal}_i, i=1 \cdots n$. The edges shorter than $50$ pixels and the cluttered interior edges are filtered out to improve the edge quality. One example of filtered thermal edge map is shown in Fig.~\ref{fig:thermal_edge_trac} (middle).

With the principle that there should be correspondences between stereo, laser, and thermal edges, we project each stereo and laser edge point to the synchronized thermal image and take the closest thermal edge as its correspondence. However, since individual stereo/laser edge point to thermal edge pixel correspondence is unstable and with many outliers, inspired by Levinson and Thrun's work \cite{c5}, we build thermal edge attraction field maps to find the correspondences globally. The attraction field maps, written as $G_i, i=1 \cdots n$, are generated by applying distance transform on thermal edge maps. According to the properties of distance transform, each pixel in $G_i$ stores the L2 distance from itself to the nearest edge. Therefore, the reversed gradient of attraction field map points to the nearest edge, which implicitly indicates the edge correspondences and ``attracts" the projected point to achieve a better edge alignment. One example of attraction field map is shown in Fig.~\ref{fig:thermal_edge_trac} (right).


\subsubsection{Edge alignment optimization} \label{eao}

In this step, we calibrate thermal extrinsic by minimizing the reprojection error of aligned edges (REAE). We first define two projection functions, which project points in stereo point clouds and laser frames onto thermal images, respectively,
\begin{subequations}
\begin{align}
    &proj^{stereo}(p) = \left[\begin{matrix} u \\ v \\ 1 \end{matrix}\right]_{p} = \frac{1}{Z_p} K^{thermal} \left( T^{-1}_{ST} \left[\begin{matrix} x_p \\ y_p \\ z_p \\ 1 \end{matrix}\right] \right)_{1:3} \label{Ya} \\
    &proj^{laser}(q) = \left[\begin{matrix} u \\ v \\ 1 \end{matrix}\right]_{q} = \frac{1}{Z_q} K^{thermal} \left( T^{-1}_{ST}\,T_{SL}\left[
    \begin{matrix} x_q \\ y_q \\ z_q \\ 1\end{matrix}\right] \right)_{1:3} \label{Yb}
\end{align}
\end{subequations}
where the laser extrinsic $T_{SL}$ has been estimated in laser calibration (Section \ref{lasercalib}); the thermal extrinsic $T_{ST}$ is the optimization target with a rough initial guess. Then, we define the cost function REAE as
\begin{equation}
\begin{aligned}
    REAE=& \sum_{i=1}^{n} \sum_{p\in \widehat{E}^{stereo}_i} G_i\left( proj^{stereo}(p) \right) \\
    +& \sum_{i=1}^{n} \sum_{q\in \widehat{E}^{laser}_i} G_i\left( proj^{laser}(q) \right)
\end{aligned}
\label{equ:REAE}
\end{equation}
where $G_i$ is the attraction field map defined in Section \ref{afm}, $\widehat{E}^{stereo}_i$ and $\widehat{E}^{laser}_i$ are sets of inlier edge points. The inliers are selected from the stereo and laser edge points based on the distance of their projections to the nearest thermal edge.
\begin{subequations}
\begin{align}
    \widehat{E}^{stereo}_i &= \{\; p\in E^{stereo}_i \;|\; G_i\left( proj^{stereo}(p) \right) \leq th \;\} \label{Za} \\
    \widehat{E}^{laser}_i &= \{\; q\in E^{laser}_i \;|\; G_i\left( proj^{laser}(q) \right) \leq th \;\} \label{Zb}
\end{align}
\label{equ:inlier}
\end{subequations}
where $th$ is the distance threshold.

We can transform the optimization problem to an unconstrained form with Lie algebra. For a stereo edge point $p\in \widehat{E}^{stereo}_i$, we define $p^{\prime}=R_{ST}p+t_{ST}=\left[X\;Y\;Z\right]^{T}$, $u=proj^{stereo}(p)$, the Lie algebra of the thermal extrinsic $T_{ST}$ is $\xi$ and its left perturbation is $\delta \xi$. Then the Jacobian matrix of the REAE cost is
\begin{equation} \label{jacob}
\frac{\partial REAE}{\partial T_{ST}} = \frac{\partial G_i}{\partial u} \frac{\partial u}{\partial p^{\prime}} \frac{\partial p^{\prime}}{\partial \delta\xi} \delta\xi
\end{equation}
where $\frac{\partial G_i}{\partial u}$ is the image gradient of $G_i$ at $u$, and the rest part is the standard 3-D to 2-D projection model \cite{yu2020line}, 
\begin{equation}
\small{
\setlength{\arraycolsep}{3.5pt}
\frac{\partial u}{\partial p^{\prime}} \frac{\partial p^{\prime}}{\partial \delta\xi} = \left[\begin{matrix} 
\frac{f_x}{Z} & 0 & -\frac{f_x X}{Z^2} & -\frac{f_x XY}{Z^2} & f_x+\frac{f_x X^2}{Z^2} & -\frac{f_x Y}{Z} \\
0 & \frac{f_y}{Z} & -\frac{f_y Y}{Z^2} & -f_y-\frac{f_y Y^2}{Z^2} & \frac{f_y XY}{Z^2} & \frac{f_y X}{Z}
\end{matrix}\right]
}
\end{equation}
The Jacobian matrix for laser edge point $q\in \widehat{E}^{laser}_i$ is similar as Eq.~\ref{jacob}.

Although the cost function is not globally convex, in reality it is always locally convex around the correct calibration \cite{c5}, thus we can use Ceres solver \cite{cit:ceres} to optimize it with a rough initialization.


\subsubsection{Rough calibration}

The proposed optimization-based method is designed for achieving high accuracy calibration with a rough initialization. To further improve the initialization error tolerance capability, we introduced a rough calibration procedure before the optimization. It consists of two consecutive grid searches on rotation (grid size $\ang{1}$) and translation (grid size $4cm$) respectively, over a given range around the initial guess of the thermal extrinsic, and pick the transformation which maximizes the number of inlier points, $|\widehat{E}^{stereo}_i| + |\widehat{E}^{laser}_i|$. The rough calibration result is then delivered to the optimizer as the initial value. With the rough calibration, our framework can work properly under large initialization errors. The user only needs to provide a rough initial guess of the thermal extrinsic, which can be obtained with ruler and protractor, or even by visual estimation.

\section{Experimental Results}

In this section, we present the evaluation of our calibration framework on real-world datasets. We first introduce our sensor setups and collected datasets, then evaluate the performance of the laser extrinsic calibration and thermal extrinsic calibration, respectively.

\begin{figure}[htbp]
    \centering
    \includegraphics[width=\linewidth]{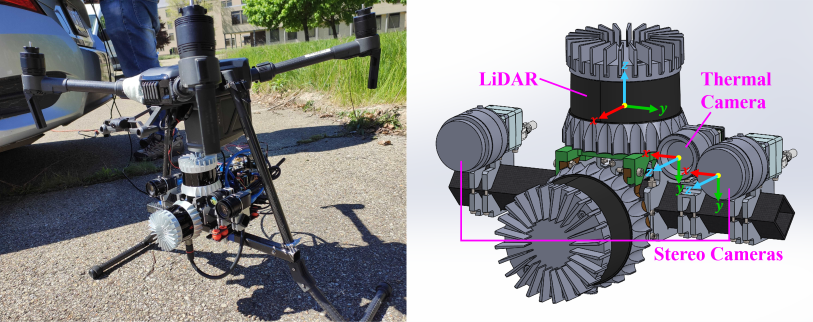}
    \caption{\emph{Left:} The entire UAV platform. The sensors are mounted under the drone. \emph{Right:} The installation position and orientation of the stereo cameras, thermal camera, and laser sensors. Notice that there are two laser sensors on our platform, but we only use the horizontally installed one.}
    \label{fig:sensors}
\end{figure}

\begin{figure*}[htbp]
    \centering
    \includegraphics[width=\linewidth]{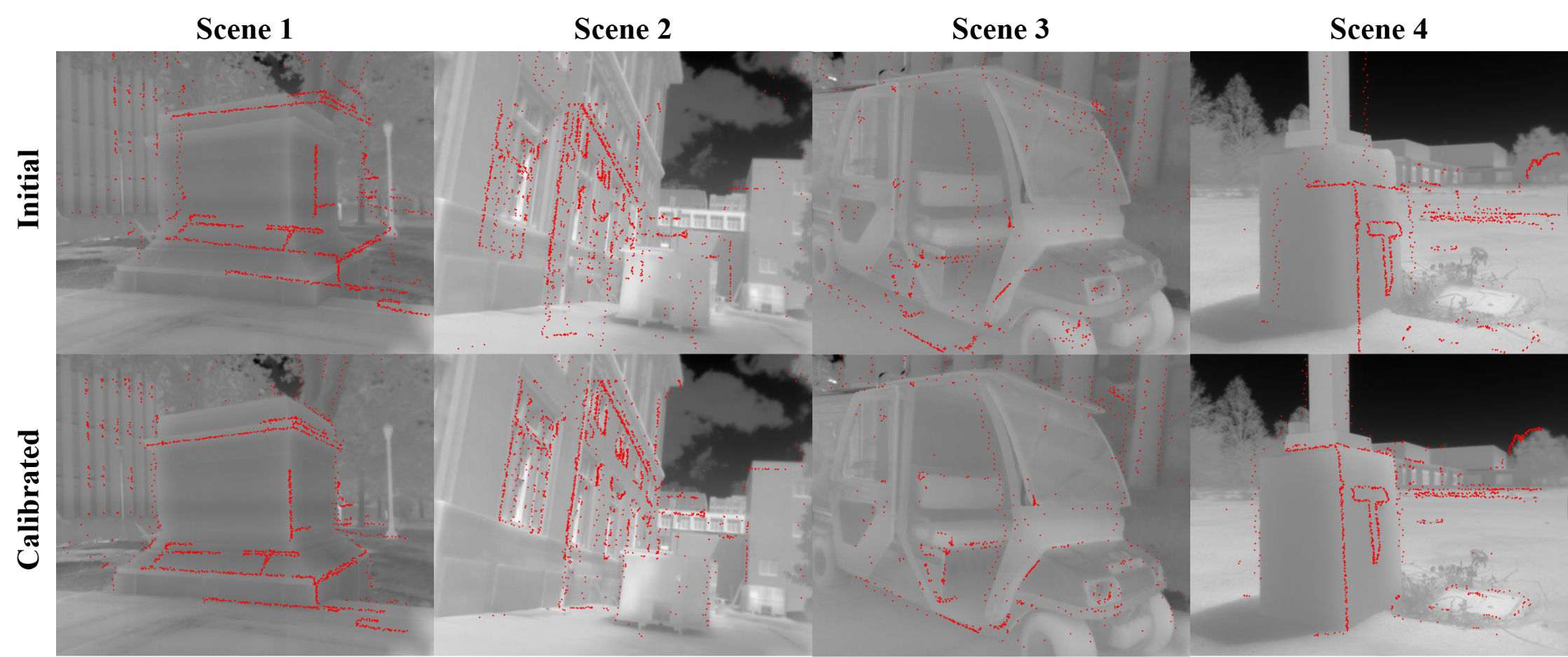}
    \caption{Stereo and laser edge projection (red dots) on thermal images before and after thermal extrinsic calibration on the 4 datasets. \emph{First row:} Projection with rough initial extrinsic; \emph{Second row:} Projection with our calibrated extrinsic.}
    \label{fig:thermal_res}
\end{figure*}

\subsection{Sensor setups and data collection}

The sensors we use include two Ximea MC124CG-SY RGB cameras, one FLIR Boson® 640 Longwave Infrared (LWIR) thermal camera, and one Ouster OS0-128 laser sensor. The two RGB cameras are fixed on both sides of the laser sensor to form a stereo pair with $baseline=22.270cm$; the thermal camera is installed between the two RGB cameras, next to the left camera; and the laser sensor is installed a little higher in the middle. All the sensors are hardware-synchronized by a PPS trigger. The stereo cameras, laser, and thermal camera are running at 2Hz, 10Hz, and 50Hz, respectively. The sensor group is mounted under a UAV platform to move freely in space.  Fig.~\ref{fig:sensors} shows the overall appearance of the UAV platform and the sensor installation schematic diagram. The stereo and thermal cameras' intrinsics $K^{left}, K^{right}, K^{thermal}$, and their distortion models have been calibrated beforehand.

We collect 4 sets of data on the CMU campus. Each data set contains synchronized stereo RGB, thermal, and laser data. Fig.~\ref{fig:thermal_res} shows the 4 scenes with one thermal frame, which contain both certain objects and background structures. 

\subsection{Laser extrinsic calibration results}

In this section, we test the performance of our framework in calibrating the laser extrinsic on the 4 datasets. To give a quantitative comparison, we manually calibrate a laser extrinsic ground truth. It is obtained by a PnP solver on manually matched laser points and RGB image pixels over 10 different frames.

\begin{figure}[htbp]
    \centering
    \includegraphics[width=\linewidth]{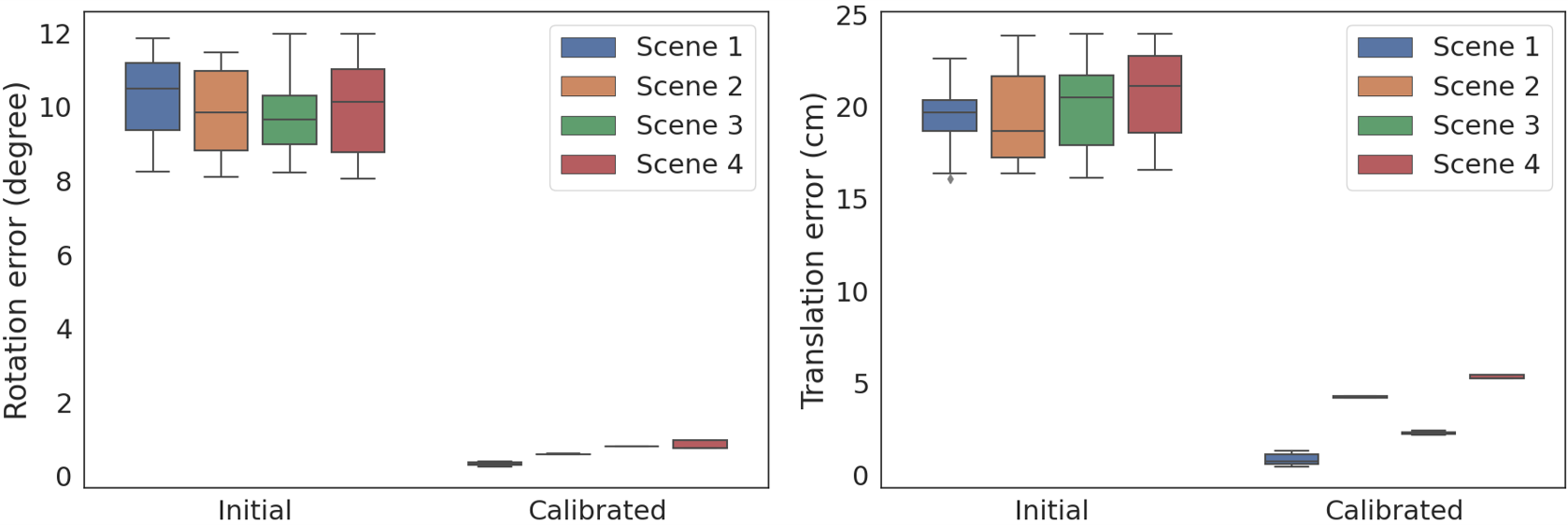}
    \caption{Laser extrinsic rotation and translation errors of the rough initial extrinsic and our calibrated extrinsic in all scene settings.}
    \label{fig:laser_err}
\end{figure}

\begin{figure}[htbp]
    \centering
    \includegraphics[width=\linewidth]{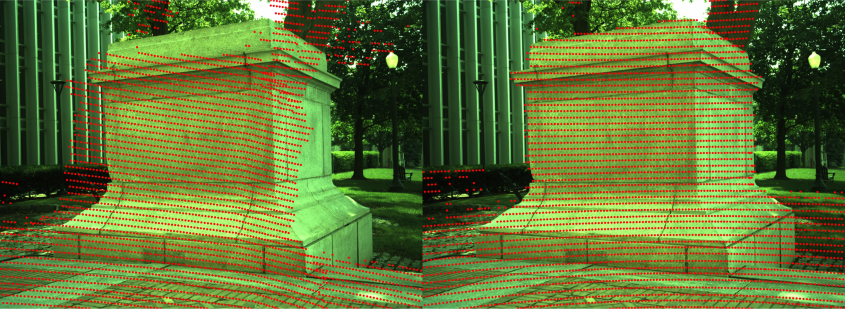}
    \caption{Laser near points projection (red dots) on RGB image with initial (left) and calibrated laser extrinsic (right).}
    \label{fig:laser_res}
\end{figure}

To validate the robustness of our laser calibration algorithm under different initialization values, we randomly sample 20 initial guesses around the ground truth (offset by $\ang{8}-\ang{12}$ for rotation, $16-24cm$ for translation). We conduct our laser calibration algorithm with each of the 20 initialization values on the 4 datasets respectively, totaling 80 tests. The extrinsic errors before and after calibration are shown in Fig.~\ref{fig:laser_err}. It can be observed that both rotation and translation errors decrease significantly after the calibration, and converge to almost the same value in each scene regardless of the initial value distribution. The final rotation errors are less than $\ang{1}$ and the translation errors are less than $5cm$ among all datasets. Fig.~\ref{fig:laser_res} shows the projection of laser near points ($depth<10m$) on RGB images with initial and calibrated extrinsic parameters. By observing the alignment of the projected points and the outline of the stone pier in the image, it can be seen that the calibration is accurate. These results suggest that our laser extrinsic calibration algorithm has good performance even with large initialization errors.

\subsection{Thermal extrinsic calibration result}

In this section, we validate our thermal extrinsic calibration algorithm in terms of cost convergence, calibration accuracy, and comparison with the checkerboard-based method. The thermal extrinsic ground truth is obtained by solving a PnP problem on the manually picked triangulated stereo points (actually from stereo images) and corresponding thermal pixels. Like the laser calibration validation, we randomly bias the ground truth to get 20 extrinsic initialization values for each dataset with rotation error range at $\ang{4}-\ang{6}$ and translation error range at $8-12cm$. Thus totally we have 80 tests with different rough extrinsic initialization values.

\subsubsection{Cost convergence validation}

\begin{figure}[htbp]
    \centering
    \includegraphics[width=\linewidth]{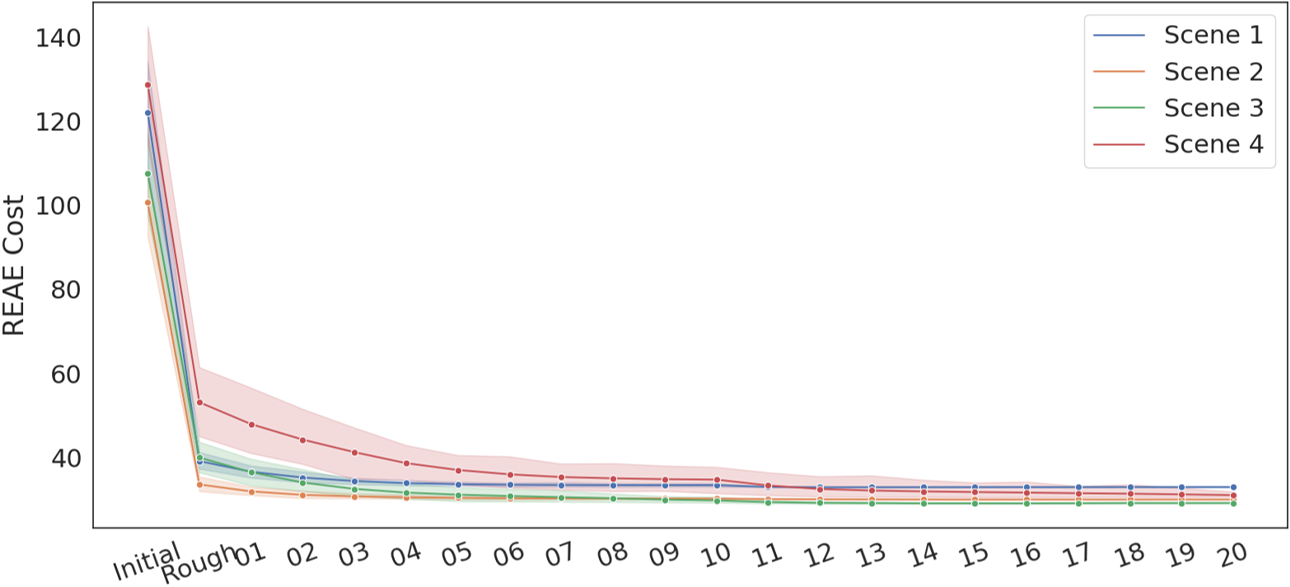}
    \caption{REAE costs over the number of iterations in all scene settings. The solid lines and dots are the average costs of all 20 tests in each scene while the shaded areas indicate the standard deviation.}
    \label{fig:costs}
\end{figure}

To verify the convergence of the cost function, we draw the cost values over the number of iterations in the 4 datasets with variant initialization values, as shown in Fig.~\ref{fig:costs}. As can be seen, the cost values gradually decrease with the number of iterations, and the variances also become smaller. All tests are sufficiently converged after 20 iterations. This demonstrates the effectiveness and robustness of our thermal calibration algorithm.

\subsubsection{Calibration accuracy validation}

\begin{figure}[htbp]
    \centering
    \includegraphics[width=\linewidth]{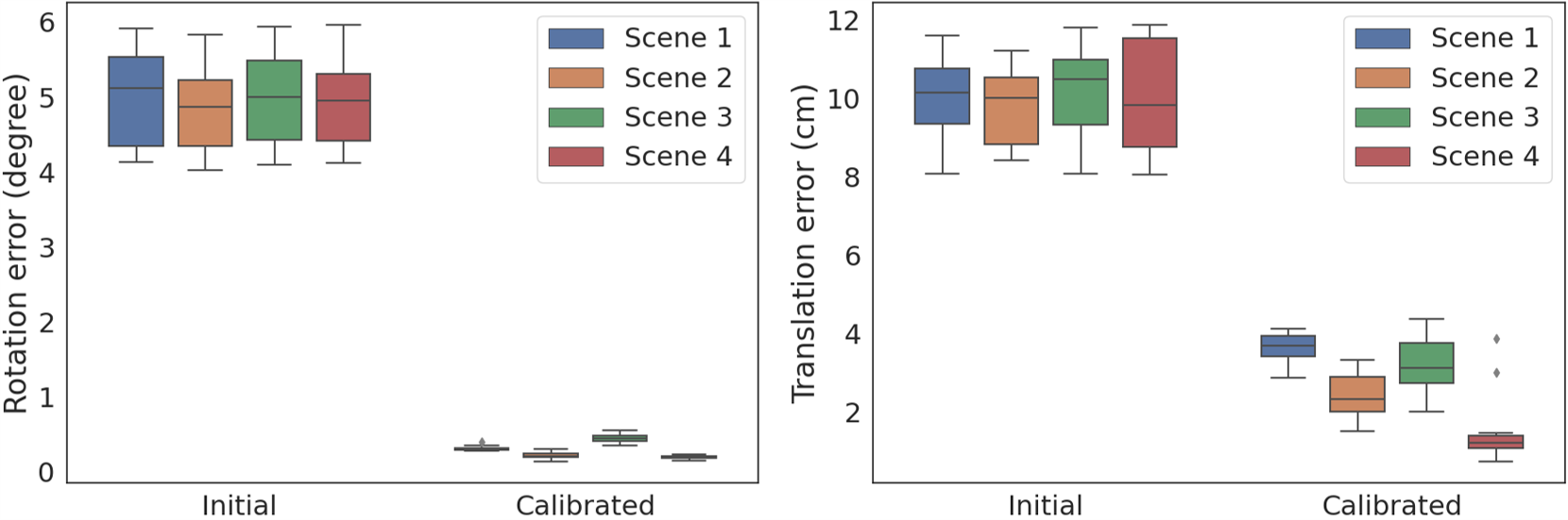}
    \caption{Thermal extrinsic rotation and translation error of the rough initial extrinsic and our calibrated extrinsic in all scene settings.}
    \label{fig:thermal_err}
\end{figure}

To quantitatively evaluate the accuracy, we calculate the extrinsic errors of the 80 tests relative to the ground truth, as shown in Fig.~\ref{fig:thermal_err}. It is clear that both the rotation and translation errors are decreased after calibration, and the calibration results converge to almost the same value in each scene, regardless of the large variance of the initialization values. The final rotation errors are less than $\ang{0.5}$ and the translation errors are less than $4cm$ among all tests. Edge points projection with the initial and calibrated extrinsic are shown in Fig.~\ref{fig:thermal_res}. It is clear that the stereo, laser, and thermal edges align well after the calibration.

\subsubsection{Comparisons}

To further demonstrate the effectiveness of our proposed method, we compare it with a traditional checkerboard-based approach. Follow the method described in \cite{c9}, we make a special checkerboard that is distinguishable in both RGB and thermal images. We use OpenCV's \cite{cit:opencv} checkerboard detector to detect checkerboard corner points in stereo and thermal images, match them according to their geometric positions, and solve a PnP problem on triangulated stereo corners and corresponding thermal corners. We also test our method with a visual estimation initialization on this dataset. Table.~\ref{tab1} lists the comparison results of the calibration errors. It can be seen that our method achieves similar accuracy to the checkerboard-based one without using a checkerboard, which demonstrates the flexibility of our proposed method.

\begin{table}[H]
\caption{Thermal extrinsic calibration comparison}
\label{tab1}
\centering
\begin{tabular}{|c|c|c|}
\hline
	 &\textbf{Our Method} &\textbf{Checkerboard~\cite{c9}} \\
\hline
	\textbf{Rot. Err. ($^{\circ}$)} &0.15 &0.17 \\ \hline
	\textbf{Trans. Err. (cm)} &3.24 &2.30 \\
\hline
\end{tabular}
\vspace{-0.2cm}
\end{table}






\section{CONCLUSIONS} 

This paper presents a multi-modal extrinsic calibration framework among stereo cameras, thermal cameras, and laser sensors. We first use multi-frame ICP to calibrate the extrinsic between stereo cameras and laser sensors, then project registered stereo and laser edge points onto thermal images and optimize the edge alignment to estimate the thermal extrinsic. Our framework can work automatically in arbitrary environments without relying on specific calibration targets. Since the system is robust to large initialization errors, the initial guess can be simply obtained by rulers, protractor measurement or visual estimation, which makes our framework ease to use. The system greatly reduces the complexity of the calibration process while achieving the same level of calibration accuracy as traditional methods.

Future works will focus on exploiting the real-time performance potential of the system by pre-filtering irrelevant information and improving the parallelism capability. The real-time calibrator can be used to continuously detect and correct the extrinsic parameters’ changes due to the sensor drift or time misalignment. Besides, more kinds of edge information, such as the edges in laser reflectivity maps, can be integrated while optimizing cloud-image edge alignment to improve the system’s accuracy and robustness.

\section{Acknowledgement}
The authors acknowledge the sponsorship of this work from the Shimizu Institute of Technology (Tokyo). 

\nocite{*}  
\bibliographystyle{IEEEtran}
\bibliography{./IEEEfull,refs}

\end{document}